# A New Theoretical and Technological System of Imprecise-Information Processing

Shiyou Lian

Xi'an Shiyou University, Xi'an, China
sylian@xsyu.edu.cn

**Abstract** Imprecise-information processing will play an indispensable role in intelligent systems, especially in the anthropomorphic intelligent systems (as intelligent robots). A new theoretical and technological system of imprecise-information processing has been founded in *Principles of Imprecise-Information Processing: A New Theoretical and Technological System*[1] which is different from fuzzy technology. The system has clear hierarchy and rigorous structure, which results from the formation principle of imprecise information and has solid mathematical and logical bases, and which has many advantages beyond fuzzy technology. The system provides a technological platform for relevant applications and lays a theoretical foundation for further research.

## 1. Genesis

Imprecise-information processing is an important component of artificial intelligence (including computational intelligence). With the development of information and intelligence sciences and technologies as well as the rise in social requirements, imprecise-information processing is becoming more and more important and urgent, and it will play an indispensable role in intelligent systems, especially in the anthropomorphic intelligent systems.

However, although people presented many theories and methods for imprecise-information processing (for instance, Fuzzy Technology), a theoretical and technological system had not yet been formed that is widely approved and has solid foundation of mathematics and logic like that for uncertain-information processing. In fact, since Zadeh proposed the concept of fuzzy sets in 1965, the fuzzy-information processing technology based on fuzzy set theory has developed rapidly and made some achievements. However, so far, some important theoretical and technical problems in fuzzy-information processing have not been solved very well, such as the shape of the membership function of a fuzzy set, the objective basis and the logic theoretical basis of fuzzy logic operators, the logical basis of fuzzy reasoning, and "the dilemma between interpretability and precision" neural-fuzzy systems encountered[2]. For this reason, not a few scholars work to improve and develop fuzzy set theory, and presented many new ideas, theories and methods, which all have their respective angles of views and characteristics. But on the whole, people haven't yet reached a common view, and the existed problems are neither solved really.

Recently, a new theoretical and technological system of imprecise-information processing founded in *Principles of Imprecise-Information Processing: A New Theoretical and Technological System*[1] just emerges as the time requires.

## 2. Architecture

The architecture of the theoretical and technological system of imprecise-information processing is shown in Figure 2-1. It can be seen from the figure that the system consists of seven modules and the logical relations among them are indicated by arrows. Of them, the module located in the bottom, i.e., *the formation of flexible linguistic values and their mathematical models*, is the basis of the whole system, and it is the origin of other modules. And the upper second model, i.e., *Imprecise-Problem Solving, Imprecise-Knowledge Discovery, and Anthropomorphic Intelligent Systems*, is the goal and application interface of the system, and it is supported by other modules. And the three modules located in the middle, i.e., *Approximate Reasoning with Flexible Linguistic Rules and Approximate Evaluation of Flexible Linguistic Functions, Truth-Degreed Logic and Flexible-Linguistic-Truth-Valued Logic,* and *Fundamental Theories of Flexible Sets and Flexible Linguistic Values*, are the main body and the core of the whole system. In addition, the first module located the top is the extension and the vertical module located the right is the cross. So, the hierarchy of the whole system is clear and the structure is rigorous.

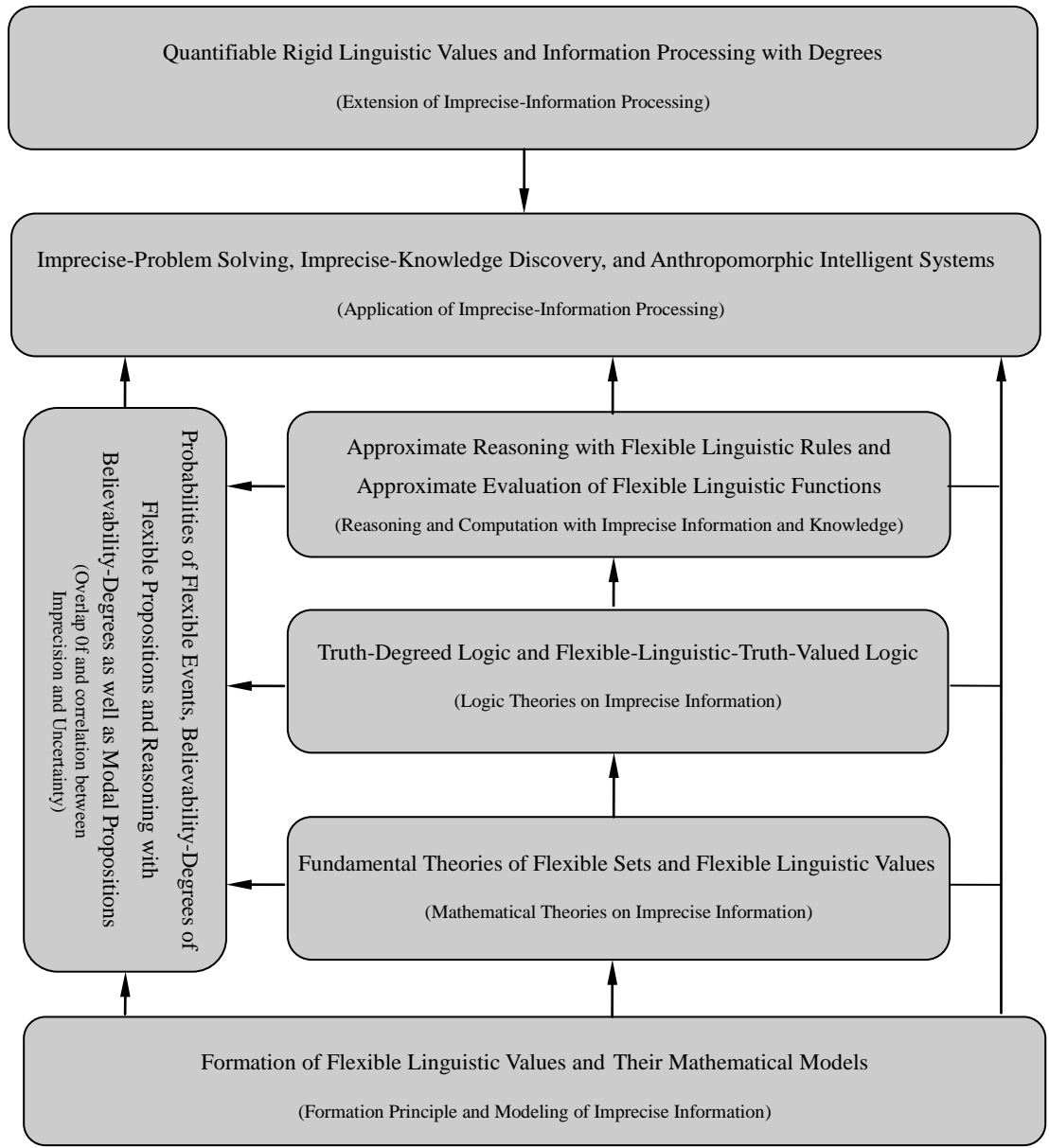

Figure 2-1 Architecture of the theoretical and technological system of imprecise-information processing

viewed from another angle, the system also has a feature, that is, in which there are many of symmetrical, antithetical or corresponding concepts and terminologies such as "flexible linguistic value" and "flexible set", "membership function" and "consistency function", "geometric model" and "algebraic model", "combined linguistic value" and "synthetic linguistic value", "form of possession" and "form of membership", "logical composition" and "algebraical composition", "conjunction-type rule" and "disjunction-type rule", "complementary flexible partition" and "exclusive flexible partition", "flexible linguistic value" and "rigid linguistic value", "medium value" and "neutral value", "L-N function" and "N-L function", "certain rule" and "uncertain rule", "natural logical semantics" and "extended logical semantics", "reasoning with truth-degrees" and "reasoning with believability-degrees", "degree-true inference" and "near-true inference", "numerical ××" and "linguistic ××", "conceptual ××" and "practical ××", "×× of single conclusion" and "×× of multiple conclusions", "×× on the same space" and "××from distinct spaces", "one-dimensional ××" and "multi-dimensional ××", "typical ××" and "non-typical ××", and so on, thus forming many pairs of parallel or complementary theories and methods —— they are arranged in a crisscross pattern, and together constitute a multidimensional system of theories and technologies.

## 3. Innovations

As shown in Figure 2-1, the theoretical and technological system is founded on the basis of formation principle of imprecise information, and from [1], imprecise information originates from the continuity or uniform chain similarity of things (see Figure 3-1). These are all revealed and proposed in [1] firstly, so the whole system is new naturally. Specifically, the system has mainly the following innovations:

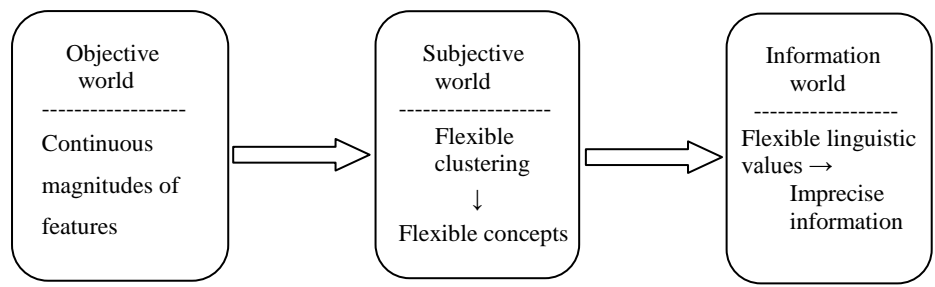

Figure 3-1 The diagram of the origin of imprecise information

(1) Treats explicitly imprecise-information processing as an independent research field, discriminates between vagueness (fuzziness) and flexibleness of concepts, proposes the terminologies of "flexible concepts" and "flexible linguistic values", and rectifies the so-called "vague (fuzzy) concepts" as "flexible concepts".

(2) Presents their general mathematical models and modeling methods according to the revealed formation principles of flexible concepts and flexible linguistic values.

(3) Proposes the concepts of flexible sets and flexible relations, and founds relevant theories and methods.

(4) Founds relevant theories and methods of flexible linguistic values.

(5) Proposes the concepts of flexible linguistic function and correlation and the concepts of flexible number and flexible function, and founds corresponding theories and methods.

(6) Introduces truth-degrees, founds the basic theory of truth-degreed logic, and finds and presents the principles and methods of corresponding inference.

(7) Introduces flexible linguistic truth values, founds the basic theory of flexible-linguistic-truth-valued logic, and presents the principles and methods of the corresponding inference.

(8) Proposes the terminology and principle of logical semantics of propositions, and based on which establishes the computation models of truth values of basic compound propositions in two-valued logic and truth-degreed logic. In addition, proposes the concepts of negative-type logic and opposite-type logic.

(9) Introduces flexible linguistic rules and founds relevant theories and methods.

(10) Introduces the adjoint functions of a flexible rule, and gives some acquiring methods and reference models.

(11) Clarifies logical and mathematical principles of the reasoning and computation with flexible linguistic rules, and gives a series of reasoning and computation approaches.

(12) Presents some approaches and ideas of the approximate evaluation of flexible linguistic functions according to the basic principles revealed.

(13) Presents the techniques and methods of problem solving according to the practical problems involving imprecise-information processing.

(14) Presents some methods and ideas of imprecise-knowledge discovery.

(15) Introduces several measures to sets and flexible sets, and founds relevant theories.

(16) Introduces quantifiable rigid linguistic values and gives relevant theories.

(17) Introduces the methodology of imprecise-information processing, and cites several application topics.

(18) Introduces random flexible events, and founds relevant probability theory.

(19) Founds the believability-degree theory of flexible propositions and presents the corresponding principle and method of reasoning with believability-degrees.

(20) Clears up the correlation between uncertain information (processing) and imprecise information (processing).

The above (1) and (2) are the basic principles of imprecise information; (3), (4) and (5) are the mathematical basic theory on imprecise information; (6), (7) and (8) are the logic basic theory on imprecise information; (9), (10), (11) and (12) are principles and techniques of inference and computation with imprecise information; (13) is the application techniques of imprecise-information processing; (14) is the acquiring techniques of imprecise knowledge; (15), (16) and (17) extend the extent of imprecise-information processing and give the basic methods and techniques of imprecise-information processing; (18), (19) and (20) are the overlapping theories and techniques of imprecise-information processing and uncertain-information processing, and clear up the connections and relations between the two.

It can be seen from above stated that the system is a new theoretical and technological system of imprecise-information processing which is different from fuzzy technology indeed. Of course, as viewed from the relationship between flexible sets and fuzzy sets, this system can also be viewed as an "amendment" to fuzzy technology. However, it is not obtained by following traditional thinking to make modifications and supplements in the existing framework of fuzzy set theory, but rather, by tracing to the source and opening a new path to research and explore imprecise-information processing with a new perspective and idea.

Examining from the perspective of quantity, how much new knowledge does this system contain? It is not hard to see that the amount of new knowledge is consistent to the meaning of information content called in information theory. With the aid of this term in information theory, the "information content" of the system is as follows:
- Over 100 important concepts.
- Over 40 theorems.
- More than 100 formulas, functions and rules.
- Over 70 specific methods and algorithms.

## 4. Advantages

From the above stated, it can be seen that the theoretical and technological system results from the formation principle of imprecise information and has solid mathematical and logical bases, and it avoids those problems occurring in the fuzzy technology. And from 3 above and theoretical analysis and example tests in [1], it is shown that the system has many advantages beyond fuzzy technology. In the following we only make a simple comparison of approximate reasoning and computation in this system and fuzzy reasoning and computation

(1) **Rationales**

It's well known that the basic principle of traditional fuzzy reasoning is CRI (Compositional Rule of Inference) proposed by Zadeh, that is, using the called composition of fuzzy relations to realize approximate reasoning. Specifically speaking, it is that first to represent a fuzzy rule (i.e., flexible rule we call) $A \to B$ into a fuzzy relation (i.e., flexible relation we call) $R$ (as definition $R =(A^c \times V) \cup (U \times B)$), and then viewing fuzzy set $A'$ approximate to $A$ also as a fuzzy relation and to do composition of $A'$ and $R$, then to treat the obtained result $B'$ (also a fuzzy set) as the result deduced by rule $A \to B$ and fact $A'$. This principle of inference is represented by an equation just as

$$B' = A' \circ R \tag{4.1}$$

The membership functional representation form of the equation is

$$\mu_{B'}(y) = f(\mu_R(x, y), \mu_{A'}(x)) \tag{4.1'}$$

where $\mu_R(x, y)$ is the so-called "implicational operator".

However, our research shows that rule $A \to B$ can merely be represented as a binary rigid relation, but not a binary flexible relation, or, a binary fuzzy relation (see Sects. 13.5 and 13.6 in [1]). In fact, in theory, rule $A \to B$ is equivalent to a binary rigid relation, but the binary rigid relation can not be definitely written out in general, so can only be represented by universal relation $core(A)^+ \times core(B)^+$ or $supp(A) \times supp(B)$.

Besides, we see that to obtain approximate conclusion $B'$, the rule $A \to B$ and fact $A'$ would be used of course, but the conduct of directly making the two intersect is improper perhaps. Because $A \to B$ is only a rule with linguistic values but not a linguistic function, and the fact $A'$ does not match the antecedent $A$ of $A \to B$, then, in this situation, doing directly composition of $A'$ and $A \to B$ is unreasonable no matter from the logic or mathematics point of view. No wonder CRI is not compatible with traditional *modus ponens*, despite CRI is called generalized modus ponens in fuzzy set theory.

Actually, the reasoning with one rule $A \to B$ is an inference on properties, while the reasoning with two rules $A \to B$ and $B \to C$ just is an inference on relations. In CRI, the minor premise, namely fact $A'$, is extended as a binary relation, but the binary relation is only a pseudo binary relation, which is still a monadic relation really. In addition, the composition of relations is conditional, not any two relations can be composed; as for composing two rules of binary relation, they must satisfy transitivity.

Also, viewed from the over ten kinds of implicational operators and the over ten kinds of formulas of relation composition, it can not be helped the objectivity of fuzzy reasoning arousing suspicion. In fact, from the implicational operator (presented by Zadeh)

$$\mu_R(x, y)=\max\{1-\mu_A(x), \min\{\mu_A(x), \mu_B(y)\}\} \tag{4.3}$$

and the formula of relation composition

$$\mu_{B'}(y)= \min\{\mu_{A'}(x), \max\{1-\mu_A(x), \min\{\mu_A(x),\mu_B(y)\}\}\} \tag{4.4}$$

It is not hard to see that $\min\{\mu_{A'}(x), \max\{1-\mu_A(x), \min\{\mu_A(x),\mu_B(y)\}\}\}$ is actually the truth-degree of premise $(A \rightarrow B) \wedge A'$ of reasoning, while also treating it as the truth-degree of conclusion $B'$ is then entirely artificially set.

Besides, fuzzy reasoning does not consider the orientation of linguistic value $A'$ relatively to antecedent linguistic value $A$ of rule $A \rightarrow B$, but which is closely related to the accuracy of conclusion linguistic value $B'$ and the exactness of corresponding number $y'$. Because viewed from the perspective of flexible linguistic functions, the approximate reasoning is also the approximate evaluation of a flexible linguistic function with a single pair of corresponding values.

Correspondingly, our approximate reasoning and computation approaches have logical and mathematical rationales.

In fact, the inferring part of our natural inference is just usual *modus ponens*; and our reasoning with degrees is following inference rule "truth-degree-level (degree-level)-UMP", and in which the numerical computation models and methods in which are founded naturally on the bases of the mathematical essence of flexible rules and approximate reasoning, and under the constraints of the logical semantics of rules and the inference rules of "near-true" and "rough-true". So it is logical. Just because of this, it is completely compatible with *modus ponens* in traditional logic, which is really the thinning and generalization of latter.

Our AT method is then presented on the basis of the numerical models and the approximate evaluation principle of linguistic functions in viewpoint of mathematics completely, so its rationality is obvious (for the specific demonstration see Sect.16.2.3 in [1]). As to other reasoning and computation methods (as interpolation method and approximate global function method) based on linguistic functions, since they are presented at the level of linguistic functions, so they have solid mathematical basis, and these methods are all closely related to mathematical backgrounds of practical problems, and our approximate computation with the adjoint measured functions of rules is then directly applying the approximate background functions of corresponding rules to be realized. Therefore, the rationality of these methods is also beyond doubt.

Actually, from the perspective of logic, our natural inference, reasoning with degrees and AT method are all the reasoning in the sense of near-true (truth-degree>0.5), that is, near-true inference we said (see Sect. 11.6 in [1]); while fuzzy reasoning can roughly be counted as the reasoning in the sense of degree-true (truth-degree>0), that is, degree-true inference we said (see Sect. 11.5 in [1]), and, the fuzzy reasoning with multiple rules (such as Mamdani-style inference and

Sugeno-style inference) can roughly be included in the parallel reasoning with degrees we said (see Sect.15.6 in [1]). From Chaps. 11, 12, 13, 14 and 15 in [1] we see that near-true inference is consistent with the natural logical semantics of propositions, so it tallies with inference mechanism of human brain; while degree-true inference is not consistent with the natural logical semantics of propositions (it corresponds to extended logical semantics of propositions), hence it does not tally with inference mechanism of human brain, particularly, the parallel reasoning with degrees based on the degree-true inference is more different from the inference mechanism of human brain; in addition, after parallel reasoning with degrees, for the synthesizing of multiple conclusions, there is no a unified method or model having theoretical basis.

(2) **Efficiencies and effects**

Comparing near-true inference and degree-true inference, we see that using near-true inference, one and the same problem can be solved by reasoning once, but if using degree-true inference, then it needs to be done by (parallel) reasoning generally twice or multiply. That is to say, the efficiency of our natural inference, reasoning with degrees and AT method in the sense of near-true, generally speaking, will be higher than that of fuzzy reasoning in the sense of degree-true.

Also, let us observe ranges of truth-degrees, the range of truth-degrees of near-true inference is $(0.5, \beta]$ ($\beta \geq 1$), and the range of truth-degrees of degree-true inference is $(0, \beta]$; while the range of truth-degrees of fuzzy reasoning is $[0, 1]$. Note that the 1 in $[0, 1]$ here is equivalent to $[1, \beta]$ in $(0.5, \beta]$. From the three ranges of truth-degrees, it can roughly be observed that the effect (i.e., accuracy) of our reasoning methods, generally speaking, will be better than that of fuzzy reasoning.

In the following we will further analyze the efficiencies and effects of the two kinds of reasoning.

First, fuzzy reasoning treats rule $A \to B$ as a couple or implication relation (namely, "if $A$ then $B$, else $B \vee \neg B$"). As a couple relation, the rule $A \to B$ is represented as the Cartesian product of corresponding fuzzy sets, $A \times B$, the geometric space this fuzzy set occupies in real is the region $\text{supp}(A) \times \text{supp}(B)$ in corresponding universe of discourse (as shown in Figure 4-1(a)). While in our inference methods, the rule $A \to B$ is then further represented as smaller region $\text{core}(A)^+ \times \text{core}(B)^+$ (as shown in Figure 4-1(b)).

As an implication relation, rule $A \to B$ is also the following two correspondences:
$$\begin{cases} A \mapsto B \\ \neg A \mapsto B \vee \neg B \end{cases}$$
Which is represented as region $(A^c \times V) \cup (U \times B)$ (as shown in Figure 4-1(c)) in fuzzy reasoning. Since the evidence fact $A'$ of the reasoning is an approximate value of $A$, so the above second correspondence "$\neg A \mapsto B \vee \neg B$" is really superfluous. That is to say, when rule $A \rightarrow B$ is treated as an implication relation there is redundancy in fuzzy reasoning.

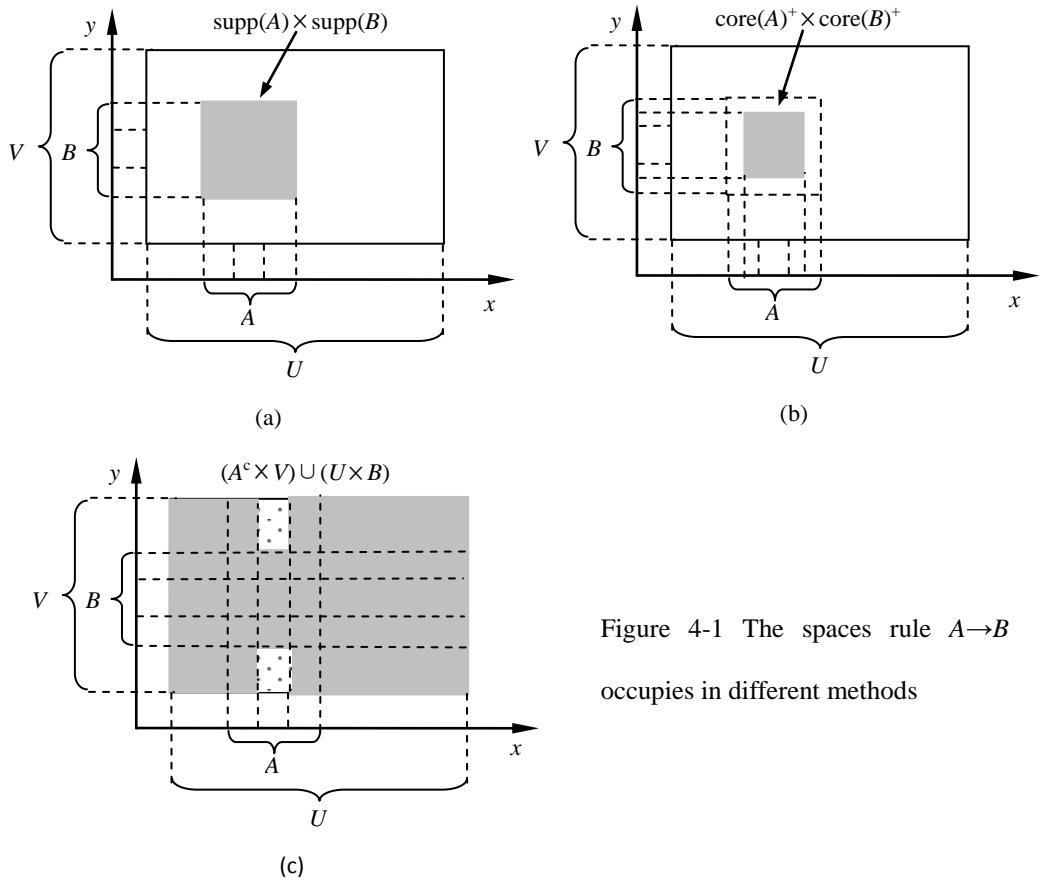

Figure 4-1 The spaces rule $A \rightarrow B$ occupies in different methods

Yet our reasoning methods only use correspondence $A \mapsto B$ in implication relation like the usual logical inference. Therefore, the reasoning can be implemented but also there is no any redundancy.

Secondly, seeing from processes, fuzzy reasoning is based on region $supp(A) \times supp(B)$ or $(A^c \times V) \cup (U \times B)$ to realize approximate reasoning and computation. From the analysis above it can be known that region $(A^c \times V) \cup (U \times B) - core(A)^+ \times core(B)^+$ is practically useless in reasoning. Then, the effect is conceivable finding the approximate value $B'$ or corresponding number $y'$ based on such a space much bigger than the actual requirement.

Yet our natural inference and reasoning with degrees realize approximate reasoning and computation in the sub-region $core(A)^+ \times core(B)^+$ of $(A^c \times V) \cup (U \times B)$; our AT method realizes approximate reasoning and computation round the peak-value point $(\xi_A, \xi_B)$ in region $core(A)^+ \times core(B)^+$. From Sect. 13.5 in [1] we known that region $core(A)^+ \times core(B)^+$ is the smallest space that includes the background function or background correlation of rule $A \rightarrow B$. Thus, viewed only from problem-solving space, the error caused by our approximate reasoning methods is certainly not more than that by fuzzy reasoning, on the whole. As to our other approximate reasoning and computation methods utilizing linguistic functions, because of their mathematical rationality, therefore their efficiency and effect are also assured.

When fuzzy reasoning is applied to approximate computation of numerical values (as auto-control), the properties of the input and output data generally are needed to be changed through fuzzification and defuzzification (as shown in Figure 4-2).

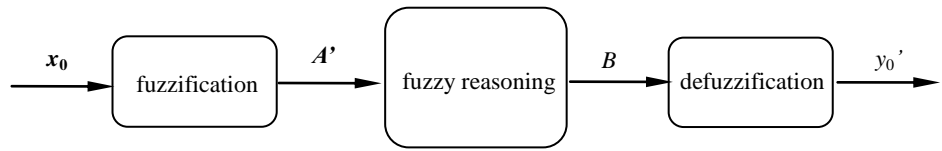

Figure 4-2 Diagram of principle of approximate evaluation computing of fuzzy controller

But our AT method can be directly applied to approximate evaluation of a function, which is tantamount to omitting two procedures of fuzzification and defuzzification. Although our approximate evaluation methods that utilize natural inference, reasoning with degrees and flexible linguistic function have also the conversion procedure from a linguistic value to a numerical value, that is, L-N conversion, our conversion can always guarantee the numerical result ($y_0'$) obtained falls in the extended core of corresponding flexible linguistic value (*B*) ( as shown in Figure 4-3(a) ), so it is always effective and satisfactory. However, the methods of fuzzy reasoning and follow-up defuzzification can not guarantee that the obtained numerical result ($y_0'$) falls within the extended core of corresponding flexible linguistic value (*B*) (as shown in Figure 4-3(b) ), as a result, there would occur the phenomenon of sometimes valid but some other times invalid as well as valid for some problems but invalid for some other problems. We think, it just is an important cause of the fuzzy control being not reliable and stable enough.

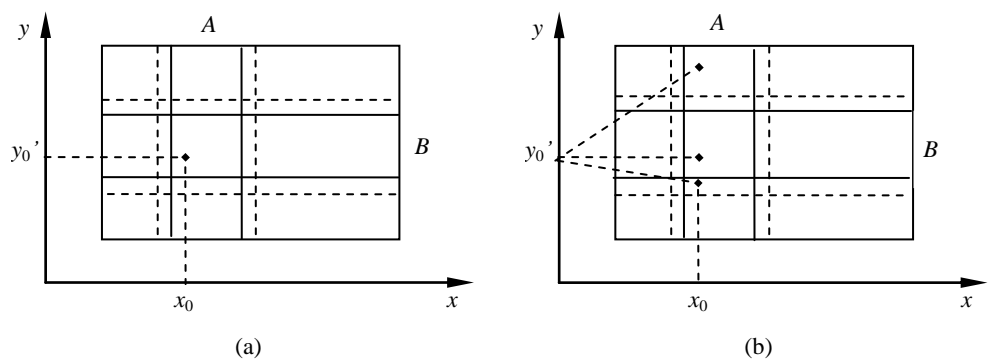

Figure 4-3 Examples of the positions of numerical result ($y_0'$) obtained by two different approximate computing models

The region between two vertical dotted lines is support set of flexible linguistic value *A*, and between two solid lines is the extended core of *A*;

The region between two horizontal dotted lines is support set of flexible linguistic value *B*, and between two solid lines is the extended core of *B*.

Actually, fuzzy controller is a kind of interpolator essentially (this is just the cause why fuzzy reasoning can be used to approximate computation to solve the practical problems such as automatic control). But viewed from approximate computation, this interpolate method is inefficient (the cause is as stated above). The approximate computing model of a fuzzy controller is considered to be able to further develop into a fuzzy logic system which can realize "universal approximator", but this kind of approximate computing systems has a severe problem —— its principle can not be clearly explained (see Sect. 16.4.3 in [1]). But the kinds of approximate computing models given in Chaps. 15, 16 and 17 in [1], only need to add respectively a function to reduce dynamically granule sizes of flexible linguistic values, then they can all realize the "universal approximator", and the functions approximated are more widespread, for example, which can also be a multiple valued function and vector function.

(3) **Research methods**

In research of fuzzy reasoning, the mathematical backgrounds of practical problems are considered rarely, the orientation of an approximate linguistic value is still less considered, but it is tried in the same way that at the level of logic and language to research a fuzzy logic method that can treat the practical problems in different poses and with different expressions.

However, since $(A \to B) \wedge A' \Rightarrow B'$ is not a valid argument form, so it is impossible to deduce *B'* directly from $A \to B$ and *A'*. On the other hand, logical relation can not reflect, or in other words, covers, the orientation relation when one linguistic value approaches to another linguistic value, but the orientation relation is of great importance to the *B'* to be obtained.

We know that a linguistic rule actually summarizes an infinite number of practical binary relations —— correlations or functions. And relative to the practical binary relation summarized by rule $A \to B$, the binary relation $core(A)^+ \times core(B)^+$ is a universal relation, while binary relation $(A^c \times V) \cup (U \times B)$ is a bigger universal relation. Therefore, the approximate reasoning at the level of linguistic values should have a guide of the lower-level numerical values, i.e., the mathematical background. Otherwise, with only logic or mathematical methods, it's difficult to obtain a desired effect. On the other hand, the approximation of flexible linguistic value *A'* to *A* also involves the orientation problem; different orientations may bring different inference results. However, in the methods of pure logical inference, the information about orientation can not be reflected, so *A'* approximates to *A* no matter from which direction, the result *B'* is all the same. Obviously, the efficiency of the result of such approximate reasoning can only go by luck.

But our approximate reasoning methods are just guided by the relevant mathematical background information of practical problems, where linguistic values only have the effect of macroscopic positioning, while the specific computations are also different with different problems.

(4) **Scopes and abilities**

The rules with multiple conditions involved in fuzzy reasoning are only two types of conjunction and disjunction, but our flexible rules have a type of synthesis-type rule besides the two types of rules. That is to say, compared with fuzzy reasoning, our methods have a more wide scope of application and more powerful processing ability.

## 5. Applications

Application is the goal of the system. Actually, the system provides a technological platform for relevant applications. In fact, using the theories, techniques and methods in the system we can directly develop related applications such as intelligent robots, expert (knowledge) systems, machine learning, natural language processing as well as various anthropomorphic computer application systems used for classifying, recognition, judging, decision-making, controlling, diagnosis, forecasting, translating, etc.

And then, the system also lays a theoretical foundation for further research. In fact, on the basis of the system we can also carry out further researches. Six research directions and some topics are given in the literature [1]:

(1) Development of anthropomorphic computer application systems and intelligent systems with imprecise-information processing ability.

(2) Imprecise-knowledge discovery, and machine learning with imprecise information. Actually, preliminary study shows that based on this theoretical and technological system, in theory, we can also achieve an alternative Deep Learning.

(3) Natural language understanding and generation with flexible concepts.

(4) Flexible logic circuits and flexible computer languages.

(5) Exploring on the brain model of flexible concepts and the qualitative thinking

mechanism of human brain.

(6) Related mathematical and logic theories.

# References

[1] Shiyou Lian. *Principles of Imprecise-Information Processing*: *A New Theoretical and Technological System*, Springer Singapore, 2016.

[2] Jyh-Shing Roger Jang, Chuen-Tsai Sun, Eiji Mizutani. 1997. *Neuro-Fuzzy and Soft Computing* (Prentice Hall, Upper Saddle River, NJ). pp: 342~245, 382~385.